\documentclass{article}

\PassOptionsToPackage{numbers, compress}{natbib}
\usepackage[preprint, final]{neurips_2026}

\usepackage[utf8]{inputenc} 
\usepackage[T1]{fontenc}    
\usepackage{hyperref}       
\usepackage{url}            
\usepackage{booktabs}       
\usepackage{amsfonts}       
\usepackage{nicefrac}       
\usepackage{microtype}      
\usepackage{xcolor}         

\usepackage{amsmath}
\usepackage{bbm}
\usepackage{graphicx}
\usepackage{tcolorbox}
\tcbset{promptbox/.style={colback=gray!10, colframe=gray!40, 
    fonttitle=\bfseries, left=4pt, right=4pt}}

\title{ROAM-GRPO: Enforcing Reasoning Diversity in GRPO via Routed Off-mode Approach Mining}
\workshoptitle{Transitioning from Pre-Training to Post-Training}

\author{%
  Jeff Duan\\
  Caltech\\
  \texttt{jxduan@caltech.edu} \\
  \And
  Matt Thomson \\
  Caltech \\
  Living Intelligence \\
  \texttt{mthomson@caltech.edu} \\
}

\begin{document}

\maketitle

\begin{abstract}
Supervised fine-tuning (SFT) paired with Reinforcement Learning with Verifiable Rewards (RLVR) has become the dominant recipe for post-training LLMs on complex reasoning tasks, but has been shown to collapse the policy into a low-diversity mode. Repeated LLM outputs are nearly identical, and pass@k accuracy converges or degrades as a result. We propose ROAM-GRPO (Routed Off-mode Approach Mining for GRPO), an augmentation to the GRPO algorithm which incentivizes exploration of distinct reasoning strategies and aims to increase output creativity during evaluation. ROAM-GRPO detects well-solved problems during training and transfers them to an “off-mode” track. The model is prompted to produce distinct solution strategies to such problems, which are verified for logical validity and distinctness by an LLM judge and subsequently self-distilled into the policy. 
Training Qwen3.5-2b on a 1,000-problem subset of the MATH lvl5 dataset, ROAM-GRPO implementation increases peak pass@16 by $4.7\%$ and correct embedding cosine distance by \textasciitilde 1.5-2x on a difficult evaluation set.
\end{abstract}

\section{Introduction}
\label{sec:intro}

Modern LLM post-training on complex reasoning tasks (such as competition-level mathematics and code generation) typically consists of two phases. Supervised Fine-Tuning (SFT) trains the base model to replicate curated prompt/response pairs, while Reinforcement Learning with Verifiable Rewards (RLVR) \cite{guoRlvr} assigns reward to self-discovered trajectories. This pairing is effective at improving a model’s accuracy in a single-attempt setting (pass@1) \cite{chenHumaneval}, but has been shown to concentrate policy probability density into a narrow subset of possible reasoning trajectories \cite{yuDapo} \cite{chenDragrpo}. The result is known as mode collapse: repeated queries to reasoning problems yield highly similar outputs. This narrowed output range is especially prohibitive in cases where problem-solving creativity across multiple attempts is desired, such as highly difficult mathematical problems or Lean proof generation \cite{lample2022hypertree}.

Within popular RLVR algorithms such as Group Relative Policy Optimization (GRPO) \cite{shaoDeepseek}, mode collapse is commonly attributed to a sparse reward system (such as final answer accuracy) that fails to incentivize model exploration during training, such that probability density often accumulates within the reasoning trajectories discovered earliest. As a result, accuracy across multiple attempts (pass@k)  plateaus or degrades \cite{chenHumaneval} \cite{dangCollapse}. 

Mode collapse has spurred the development of various diversity-aligned RLVR algorithms, including those centered around mathematical reasoning. Notably, Upskill \cite{shahUpskill} conditions the policy on a discrete latent variable and assigns a token-level mutual information reward, driving outputs to be identifiable by their conditioned latent. In ideal cases, different latent variables evolve to represent distinct reasoning “skills”. However, the token-level granularity lends itself to reward hacking; distinct latents can produce outputs that differ only in phrasing and formatting, or in worse cases, high-entropy “gibberish” text. DRA-GRPO \cite{chenDragrpo} opts instead to use cosine-similarity in text embeddings to reweight group completions, although reported examples once again differ largely in style and voice (e.g. "thinking out loud" vs. cautious step-by-step reasoning) without changes to the underlying method. \citet{chenDragrpo} also report exclusively pass@1 accuracy, making it unclear what effect semantic coverage has in a pass@k environment.

We propose ROAM-GRPO (Routed Off-mode Approach Mining for GRPO), a modification of GRPO that targets increased diversity at a reasoning level (rather than paraphrasing a consistent underlying method). ROAM-GRPO detects problems that are well-solved (in which group completion accuracy reaches a fixed threshold) and routes them to an “off-mode” self-distillation track. Here the model receives a modal solution and is prompted to produce a distinct approach solving the same problem. Outputs are verified for logical validity and distinctness by an external LLM judge, and accepted outputs are distilled into the policy. 
ROAM-GRPO aims to counteract probability accumulation along a single reasoning trajectory during training, and investigates whether this reasoning exploration reflects an increase in pass@k coverage.

\section{Preliminary}
\label{sec:prelim}

\subsection{Pass@k}
Pass@k accuracy measures the probability that at least one of $k$ independent model outputs solves a problem (equivalently, the complement of the probability that none of $k$ outputs solves a problem). We use the Chen unbiased estimator \cite{chenHumaneval}, where given a problem $q$ yielding $c_q$ correct completions:

\begin{equation}
  \widehat{\text{pass@}k}(q) \;=\;
  1 - \frac{\binom{G-c_q}{k}}{\binom{G}{k}},
\end{equation}

where $G$ is the group completion size and $\binom{G-c_q}{k}/\binom{G}{k}$ is taken to be
$0$ when $G-c_q < k$.

\subsection{GRPO}
Let $\pi_\theta$ be the trained policy and $q$ a training prompt. Group Relative Policy Optimization (GRPO) \cite{shaoDeepseek} samples a group of completions $\{o_i\}_{i=1}^{G} \sim \pi_\theta(\cdot \mid q)$ and assigns each a verifiable reward $r(o_i)$. In our training, we take $r(o_i)\in\{0,1\}$ to be binary final-answer correctness.
Writing $\bar r = \frac{1}{G}\sum_i r(o_i)$, each completion receives a
group-relative advantage
\begin{equation}
  \hat{A}_i \;=\; \frac{r(o_i) - \bar r}{\mathrm{std}\big(\{r(o_j)\}_{j=1}^G\big)}.
\end{equation}
The GRPO loss is then defined as
\begin{equation}
  \mathcal{L}_{GRPO}(\theta; q) \;=\;
  -\,\frac{1}{G}\sum_{i=1}^{G}\frac{1}{|o_i|}\sum_{t=1}^{|o_i|}
  \min\!\Big(\rho_{i,t}\,\hat{A}_i,\;
  \mathrm{clip}(\rho_{i,t}, 1-\epsilon, 1+\epsilon)\,\hat{A}_i\Big),
\end{equation}
where $\rho_{i,t} = \frac{\pi_\theta(o_{i,t}\mid q, o_{i,<t})}{\pi_{\theta_{\text{old}}}(o_{i,t}\mid q, o_{i,<t})}$
. Note that we use no KL penalty during our training ($\beta = 0$).

\section{Method}
\label{sec:method}

\begin{figure}[h]
  \centering
  \includegraphics[width=0.9\linewidth]{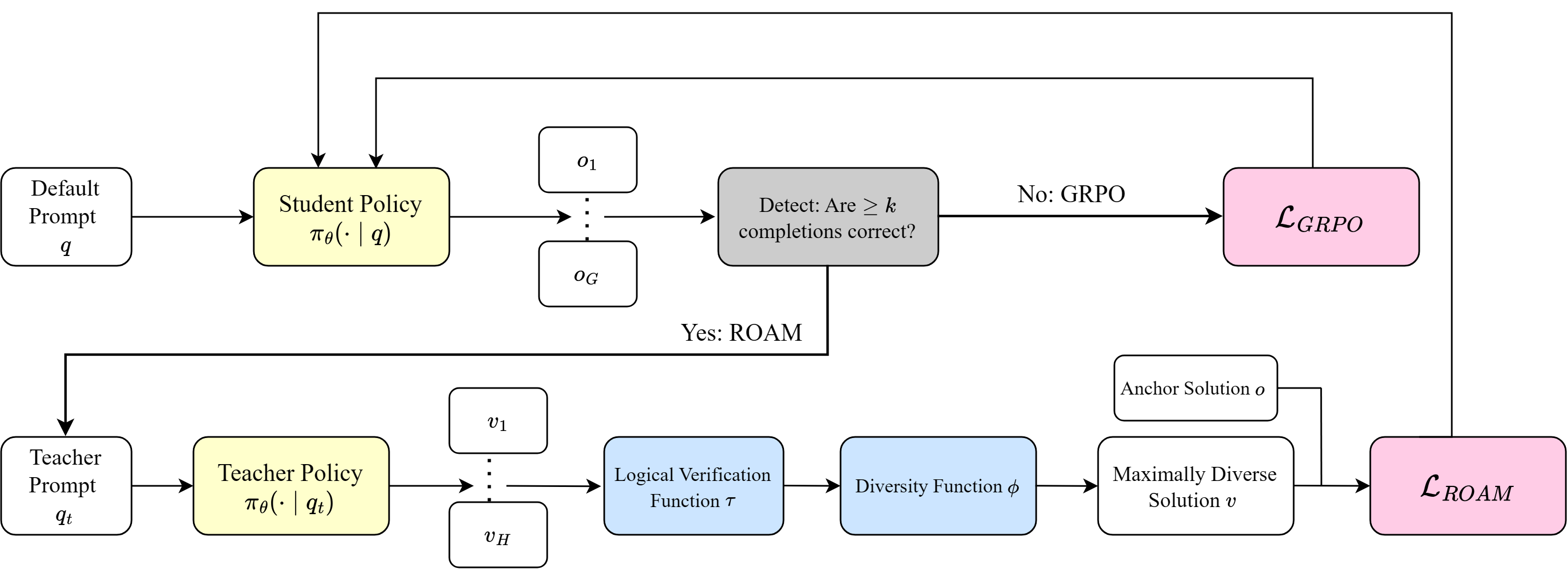}
  \caption{ROAM-GRPO training pipeline. Well-solved prompts ($\ge k$ of $G$ completions correct) are routed off the GRPO objective onto an off-mode track that self-distills valid, distinct alternative solutions measured by $\tau,\phi$ (abstract functions, instantiated as an LLM judge here in practice).}
  \label{fig:pipeline}
\end{figure}

We present a detailed description of the ROAM-GRPO training algorithm. 
It should be noted that Fig. \ref{fig:pipeline} depicts two separate, general functions $\tau, \phi$ that respectively measure an output's logical coherence and diversity. In practice, we use an LLM-judge \cite{zheng2023judging} for both purposes. The choice to delineate and generalize $\tau, \phi$ in the framework emphasizes that
the two functions may be estimated using different equipment depending on the trained task and use cases. We discuss possible alternatives in Section \ref{sec:conclusion}.

\paragraph{ROAM vs. GRPO Router.}
Given a problem $q$, the model produces a group of completions $o_1, \dots, o_G$ under standard GRPO prompting, each scored for final answer correctness. The router then assigns $q$ to either the GRPO or ROAM track 
based on a fixed threshold $k$: if $\geq k$ out of $G$ completions yield correct final answer (indicating that the policy may be converging to a replicable, consistent reasoning trajectory), $q$ is transferred to the off-mode track.

\paragraph{Off-mode Approach Mining.}
An off-mode group contains several correct solutions, although they typically share a common strategy. We therefore elicit a new approach to the problem.
A teacher completion $o_t \in \{ o_i\}_{i=1}^G$
is selected and placed within a teacher prompt $q_t$, which presents $o_t$ as "Method 1:" and instructs the policy to produce "Method 2 (a genuinely different approach)".
Sampling from the policy under this prompt produces a new pool of alternative completions $v_1, \dots, v_H$.

In practice, we select $o_t$ to be the median-length completion of the group $\{ o_i \}$, avoiding an overly terse or rambling teacher prompt. Exact details on the construction of $q_t$ are included in Appendix \ref{app:prompts}.

\paragraph{Logical Verification.}

Each off-mode completion $v_1, \dots, v_H$ is evaluated using a logical verification function $\tau$, where $\tau(v_i) = 1$ if $v_i$ is logically sound and $\tau(v_i) = 0$ otherwise. 
In practice, we instantiate $\tau$ as two separate tests for solution soundness. First, we check for symbolic correctness by extracting the candidate's final answer. Second, we impose reasoning correctness through a separate LLM judge, which is prompted to review the attempted $v_i$ and flag any instances of logical error, before emitting a final verdict of "VALIDITY: [sound|unsound]". Exact prompts and parameters are included in Appendix \ref{app:prompts}.

\paragraph{Diversity Function.}

The diversity function $\phi$ scores how distinct a candidate's reasoning strategy is from the modal solution. We likewise instantiate $\phi$ using an LLM-judge, prompted to compare a candidate $v_i$ against the teacher completion $o_t$ in terms of reasoning approach (explicitly prompted to ignore surface level variation such as voice, syntax, and verbosity) before outputting a score between $0$ (entirely distinct) and $100$ (entirely equivalent). We consider $v_i$ for self-distillation if $\phi(v_i) < 95$. 

\paragraph{Self-Distillation and ROAM Loss.}

The off-mode update distills two solutions into the policy: first, of the sufficiently distinct and logically sound completions, we choose the maximally-diverse solution $v = \arg\min_{v_i} \{\phi(v_i) \mid \tau(v_i)=1, \phi(v_i)<95\}$. Additionally, we select the shortest-length (avoiding incoherent rambling/spiraling reasoning chains) correct completion $o \in \{o_i\}_{i=1}^G$ as an "anchor solution"
to prevent the new solution from overwriting historical approaches.
Given weighting hyperparameter $\lambda$, the ROAM loss is then the standard token-level cross-entropy loss on $o, v$:
\begin{equation}
  \mathcal{L}_{ROAM}(\theta; q) \;=\;
  -\lambda (\log \pi_\theta(v\mid q)\; + \log \pi_\theta(o\mid q)).
\end{equation}

\section{Experiment and Results}
\label{sec:experiments}

\subsection{Experimental Setup}
\label{subsec:setup}

We train Qwen3.5-2b \cite{yangQwen3} with LoRA adapters ($r=8, \alpha=16$) \cite{huLora} on a 1{,}000-problem subset of the MATH train set \cite{hendrycks2021measuring}, restricted to Level-5 difficulty problems in order to avoid straightforward problems in which few reasoning approaches exist. Qwen3.5-2b was deliberately chosen over larger and more saturated models, which showed signs of containing multiple pre-memorized solutions to available datasets. Specific sampling hyperparameters are included in Appendix \ref{app:hyperparams}.

Three separate training runs are compared: (i) a \textbf{GRPO} baseline; (ii) \textbf{ROAM-GRPO} applied from the start of training; and (iii) \textbf{Delayed ROAM-GRPO}, in which ROAM is only activated starting at global step $5000$ of GRPO and starts from an existing GRPO checkpoint. 
GRPO uses $G=8$. ROAM-GRPO uses $k=G=8$, $H=16$, $\lambda=0.2$, and Qwen3.5-9b as a judge.
(The judge need not be a more capable model than the trained model in general. We use the 9b judge because it is more reliable for assessing diversity and soundness metrics than Qwen3.5-2b itself.) 

Evaluation is conducted on two held-out test sets: MATH (400 test problems of maximum (level 5) difficulty) and the more difficult Omni-MATH (400 test problems, difficulty levels 5-6) \cite{gaoOmnimath}. 
Along with pass@k accuracy, we also report mean pairwise cosine distance among the embeddings of correct completions (which we refer to as "solution dispersion"). Embeddings are generated by Qwen3-Embedding-0.6b \cite{zhangQwenembedding}. While embedding distance does not always reflect a difference in reasoning (as opposed to style), we use them as one means of visualizing the effects of ROAM.

\subsection{Training Dynamics}
\label{subsec:training}

\begin{figure}[h]
  \centering
  \includegraphics[width=0.9\linewidth]{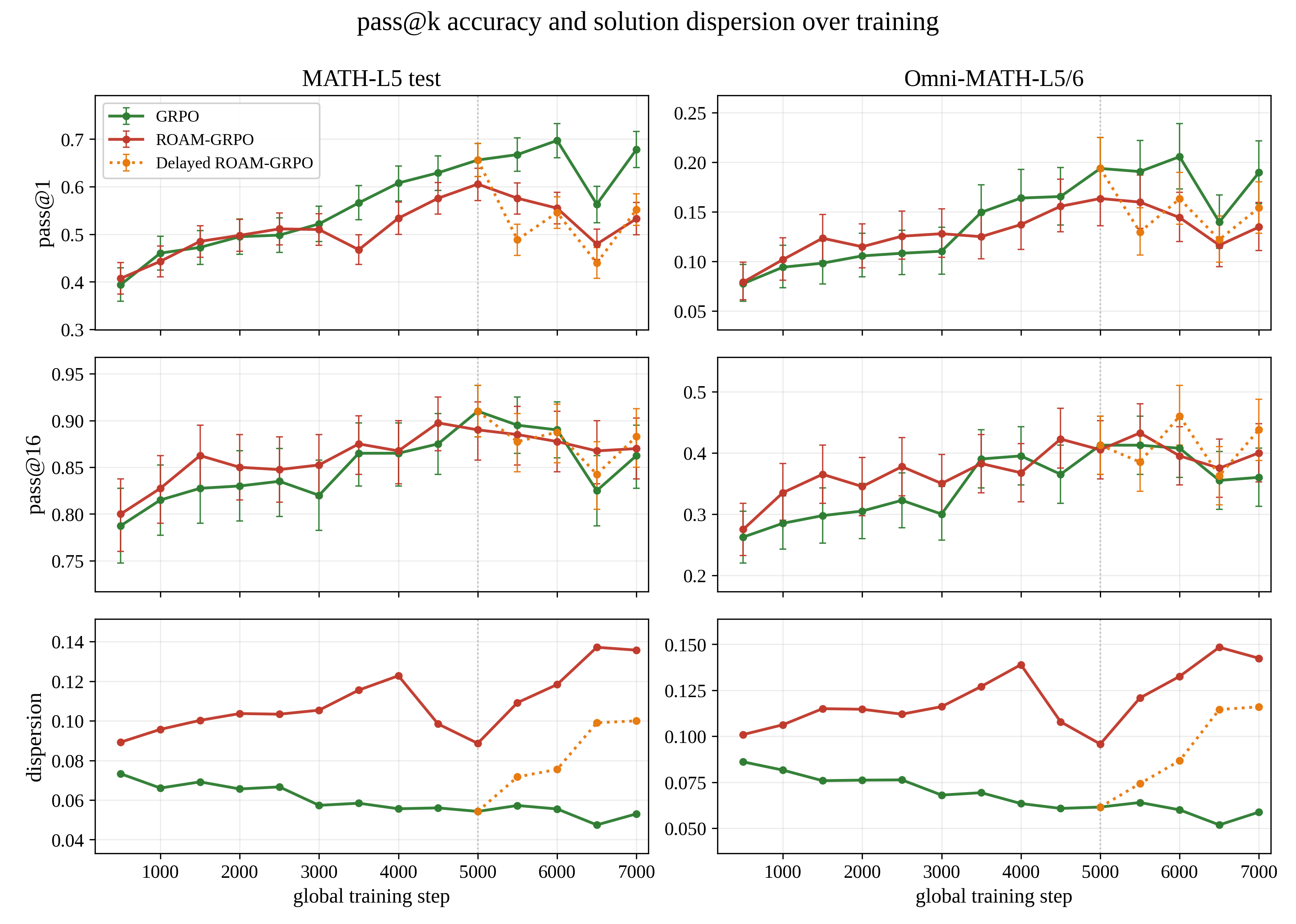}
  \caption{pass@1, pass@16, and solution dispersion over training (MATH-L5, left; Omni-MATH L5/6, right). GRPO, ROAM-GRPO, and delayed ROAM-GRPO (activated at step $5000$, dotted). Error bars on pass@1, pass@16 display 95\% percentile bootstrap confidence intervals. ROAM trades pass@1 for approximately equal pass@16 (and higher dispersion). Delayed ROAM reaches the highest peak pass@16 on Omni-MATH. pass@4, pass@8 included in App. \ref{app:more_graphs}.}
  \label{fig:training}
\end{figure}

Throughout training, ROAM-GRPO yields lower pass@1 accuracy than GRPO on both datasets (e.g. Step 6000: ROAM-GRPO yields $55.5\%$ pass@1 vs.\ GRPO's $69.7\%$ on MATH, $14.4\%$ pass@1 vs.\ $20.5\%$ on Omni-MATH). 
Despite this, ROAM-GRPO approximately matches GRPO in pass@16 accuracy across both evaluations. GRPO peaks at $91.0\%$ pass@16 (step $5000$) on MATH and $41.3\%$ (step 5000) on Omni-MATH, while ROAM-GRPO peaks at $89.8\%$ (step $4500$) and $43.2\%$ (step $5500$) on the same datasets. 
Solution dispersion is also constantly higher on both benchmarks (e.g. ROAM-GRPO maintains $\approx\!0.09$--$0.14$ vs.\ GRPO's $\approx\!0.05$--$0.07$ on MATH).

Additionally, delayed ROAM-GRPO exhibits the \textbf{highest pass@16 accuracy} across the Omni-MATH test set ($46.0\%$ at step $6000$, $4.7\%$ higher than the GRPO peak), suggesting that ROAM-GRPO can be effectively deployed after GRPO has raised baseline model performance (although training behavior is unstable, as pass@16 fluctuates in the following steps).
In general, the aptitude of delayed and undelayed ROAM-GRPO in a pass@16 environment (in spite of lowered pass@1) suggests that ROAM-GRPO checkpoints sample from a more diverse distribution of reasoning trajectories, and can sometimes reach a correct solution where GRPO cannot. 

\begin{figure}[h]
  \centering
  \includegraphics[width=0.9\linewidth]{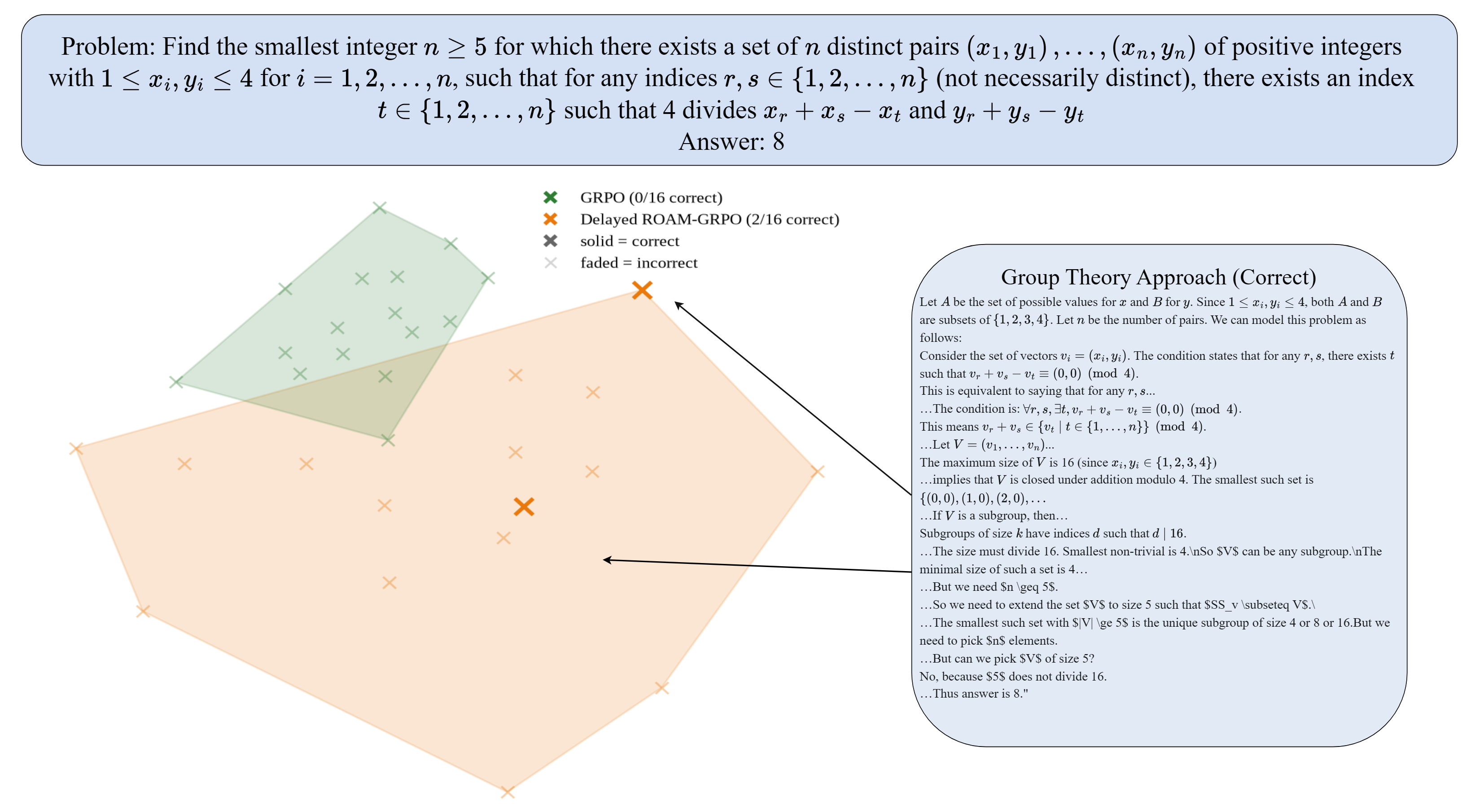}
  \caption{2-D MDS embeddings of attempted solutions on Omni-MATH Eval Problem HMMT\_2/99,
  16 each across GRPO (step 5000, green) and delayed ROAM-GRPO (step 6000, orange), taken at their highest pass@16 checkpoint. 
  GRPO's samples form a cluster and are all incorrect: roughly half attempt a group theoretic approach but wrongfully conclude answer $n=5$, while most remaining completions handwave an incorrect answer. Delayed ROAM correctly executes the group theoretic approach twice while also frequently sampling distinct approaches (residue constructions, a whole-set argument). Correct completion abridged for readability; full completions accessible in App. \ref{app:reproducibility}.}
  \label{fig:multi_embeddings}
\end{figure}

Figure \ref{fig:multi_embeddings} details one such example, along with attempted approaches. Here delayed ROAM-GRPO attempts a wider variety of approaches, and successfully executes a group-theoretic approach twice. GRPO completions also attempt a similar group-theoretic approach, but cannot produce a successful solution. 
We discuss possible interpretations in Sec.~\ref{sec:conclusion} and inspect more example problems in Appendix \ref{app:examples}, in which ROAM-GRPO similarly varies in method framing or surfaces unseen reasoning methods.

Lastly, we note that during steps $5000 \to 6000$, GRPO pass@1 increases ($65.6\%\to69.7\%$) on MATH while pass@16 falls ($91.0\%\to89.0\%$). This trend aligns with the diversity degradation documented by \citet{dangCollapse}.
The following drop at step 6500 occurs across all metrics, and thus is more likely caused by general GRPO training instabilities or overfitting.

\subsection{Method-Level Diversity Analysis}
\label{subsec:method_div}
To obtain a finer-grained view of ROAM's benefits in diversification than is offered by embeddings, we use an LLM judge (Claude Opus 5.5) to classify evaluation completions by solution method. For each held-out problem, the judge receives group completions of GRPO and of ROAM-GRPO (MATH) or delayed ROAM-GRPO (Omni-MATH) at the end of training (step $7000$). Each completion is assigned to one of a shared list of methods depending on its employed core strategy. Within each method bin, the judge further labels each pair of completions as containing "variant" (light mathematical deviations in execution, such as substitutions, reordering of steps, new casework) or "surface-level" (notation or presentation changes only) difference. This analysis is run twice: once on correct completions only, and once on all $16$ completions per arm. Specifically, only problems in which each arm produces at least one correct completion ($330$ on MATH, $130$ on Omni-MATH) are tested. The full prompt is included in App. \ref{app:prompts}.

\begin{figure}[h]
  \centering
  \includegraphics[width=0.8\linewidth]{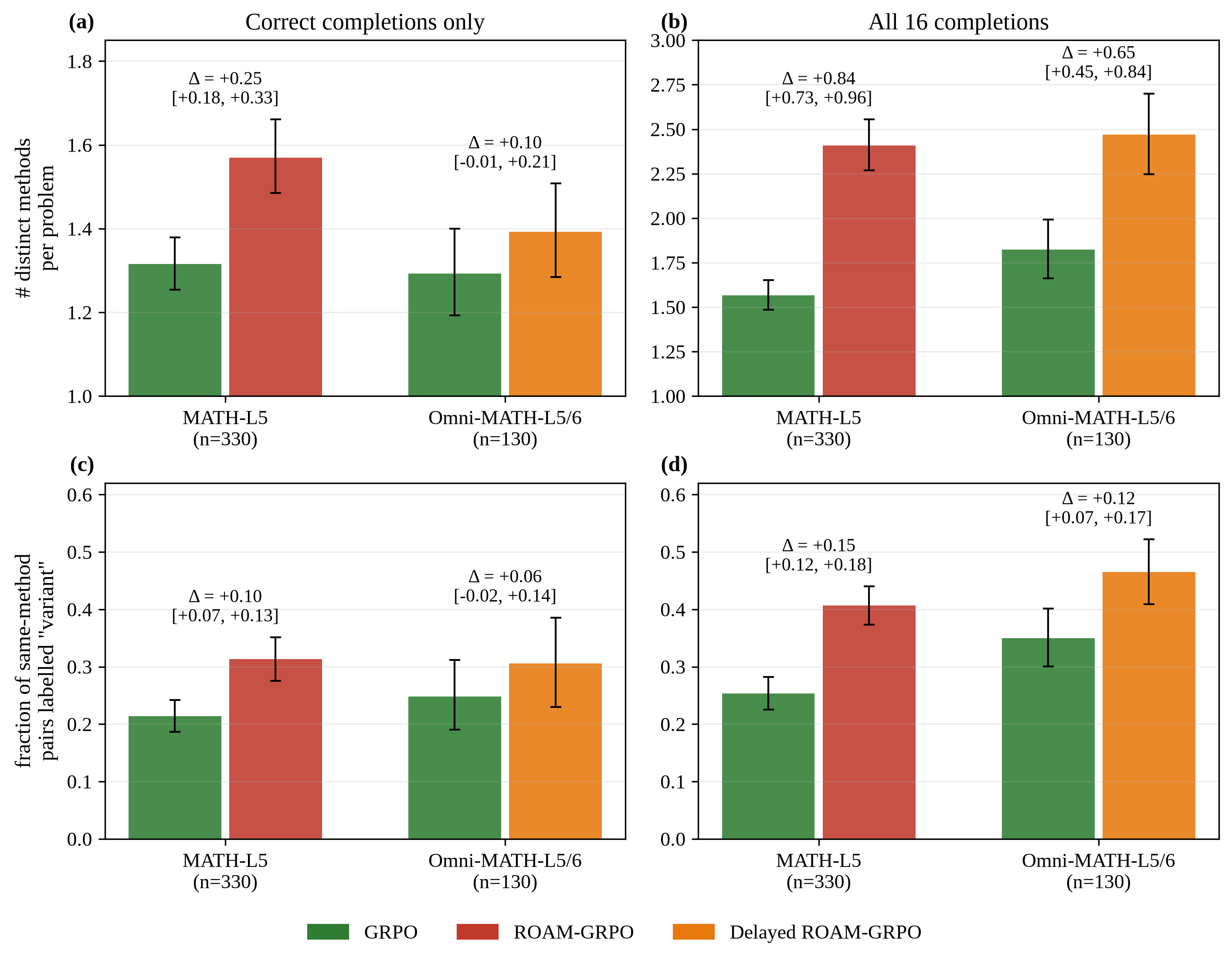}
  \caption{Method-level diversity at step $7000$ using an LLM judge (Claude Opus 5.5). Top: average number of distinct solution methods per problem, among correct completions only (\textbf{a}) and all completions (\textbf{b}). Bottom: fraction of pairs (within the same method bin) judged to be meaningful variants rather than surface level rewrites, among correct completions (\textbf{c}) and all completions (\textbf{d}). Error bars show 95\% bootstrap confidence intervals over problems, and $\Delta$ shows ROAM $-$ GRPO difference along with the 95\% CI. All judged problems require that both arms produce at least one correct completion.}
  \label{fig:method_div}
\end{figure}

As shown in Fig. \ref{fig:method_div}, ROAM produces a higher number of correct methods per problem on MATH ($\Delta=+0.25$ on correct-only and $\Delta=+0.84$ on all completions) and Omni-MATH ($\Delta = +0.10$ on correct-only and $\Delta=+0.65$ on all completions). Additionally, within the same core strategy, ROAM yields slightly more substantive mathematical variance (e.g. $\Delta=+0.10$ on MATH under correct completions). The increase in $\Delta$ under all completions compared to correct-only aligns with the observation that ROAM generates widespread diversity, at the expense of reliability in pass@1 accuracy.

\section{Conclusion} 
\label{sec:conclusion}
We developed ROAM-GRPO, an augmentation to GRPO that uses prompting and self-distillation to explore diverse reasoning paths through complex math problems. 
Notably, across Qwen3.5-2B training on MATH Level-5, delayed ROAM-GRPO increases peak pass@16 by $4.7\%$ on a more difficult set of Omni-MATH L5/6 problems, and also causes an increase in embedding distance and number of distinct employed methods. Example problems (Sec. \ref{subsec:training}, App. \ref{app:examples}) showcase ROAM-GRPO producing a more widespread set of approaches and framings.

\paragraph{Limitations.} There remain limitations to our work that point to future directions. Wider evaluation and analysis of ROAM-GRPO on larger, harder datasets would inform what aspects of mathematical reasoning ROAM most benefits. One hypothesis, suggested by  examples in Sec.~\ref{subsec:training} and App. \ref{app:examples}, is that training on a broader distribution of reasoning approaches reduces execution error on said methods when called for during evaluation, or perhaps expands the set of available reasoning methods itself.

Meanwhile, 
ROAM-GRPO training is not fully stable and does not consistently exceed GRPO in pass@16. The algorithm could benefit from further experimentation with different routing mechanisms, self-distillation implementations, and (possibly cheaper) realizations of the $\tau, \phi$ functions that mitigate decreases in pass@1 while widening pass@16 gains. Experimentation on $\tau, \phi$ could also extend ROAM-GRPO to domains where validity and structural distinctness permit deterministic definitions (such as Lean kernels in formal verification or unit tests in code generation). Applications to downstream, consecutive training frameworks remain another direction for future work.

\newpage
\bibliographystyle{plainnat}
{
\small
\bibliography{references}
}

\appendix
\raggedbottom

\section{Reproducibility}
\label{app:reproducibility}

\subsection{Setup and Code}
\label{setup}

All training runs were performed using an A100x8 cluster. Code required to reproduce the results in this paper, along with full completions for example problems, is available at the repository: 

\begin{center}
    https://github.com/jduantc/ROAM-GRPO
\end{center}

\subsection{Hyperparameters}
\label{app:hyperparams}

In addition to the core hyperparameters detailed in the main text, we use the following:

\begin{table}[h]
\caption{Sampling and judge hyperparameters. All other hyperparameters are reported in Section~\ref{sec:experiments}.}
\centering
\begin{tabular}{lc}
\toprule
\textbf{Parameter} & \textbf{Value} \\
\midrule
Learning rate & $1 \times 10^{-5}$ (constant)\\
Prompts per step & 2 \\
Max new tokens & 3072 \\
Temperature & 1.0 \\
Top-$p$ & 1.0 \\
Judge max tokens (distinctness) & 512 \\
Judge max tokens (validity) & 700 \\
\bottomrule
\end{tabular}
\label{tab:hyperparams}
\end{table}

Additionally, thinking mode is off for the trained model Qwen3.5-2b (this is the recommended post-training default on this model to avoid entering non-terminating thinking loops).

\subsection{Prompts}
\label{app:prompts}

Although they are available in the anonymized repository, we also record the specific prompts used for off-mode mining, soundness/distinctness judges, and embedding models here.

\textbf{Off-mode Mining Prompt} \mbox{}
\begin{tcolorbox}[promptbox, title=Off-mode teacher prompt $q_t$]
\textbf{[User turn]}\\
\{problem\}

Provide a solution. Put the final answer in \textbackslash boxed\{\}.

\medskip
\textbf{[Model prefill]}\\
Method 1: \{modal solution, first 300 characters\}

Method 2 (a genuinely different approach):
\end{tcolorbox}
In practice we delete whitespace and truncate $o_t$ to the first 300 characters. This allows for cheaper generation and provides a high-level excerpt of the solution without allowing the student to copy the teacher's reasoning too closely. Prompting using model prefill mitigates the use of metacommentary in the student (e.g. referencing "a different approach" in the solution text), although we also filter out student completions containing such phrases.

\textbf{Judge Prompts} \mbox{}

\begin{tcolorbox}[promptbox, title=Validity judge prompt ($\tau$)]
Audit the reasoning in the following solution. Be concise (at most 5 short lines).

PROBLEM: \{problem\}\\
SOLUTION TO AUDIT: \{candidate\}

Check specifically for:
\begin{itemize}
  \item a quantity in the problem read as the wrong kind of thing (a distance used as a time, etc.)
  \item a wrong or misapplied formula
  \item unit / dimension errors (multiplying where it should divide)
  \item arithmetic slips
  \item a conclusion that does not follow from the steps shown
\end{itemize}

Do not solve the problem yourself. Judge only whether the steps shown are logically and dimensionally valid.

End with exactly this line and nothing after it:\\
VALIDITY: sound \quad \emph{(or)} \quad VALIDITY: unsound
\end{tcolorbox}

\begin{tcolorbox}[promptbox, title=Distinctness judge prompt ($\phi$)]
You are given two solutions to the same mathematics problem.

METHOD COMPARISON.\\
STEPS A: list solution A's key mathematical moves as a numbered list of 3--6 short phrases. Name the OPERATION performed (e.g.\ `set up coordinates', `apply Vieta's formulas', `complementary counting'), not the numbers involved.\\
STEPS B: do the same for solution B.\\
VERDICT: compare the two lists. They are the SAME method if they reach the answer by the same route, even if worded differently, of very different length, or containing arithmetic slips. They are DIFFERENT only if at least one step differs in KIND: a different theorem or identity invoked, a different representation of the object, casework versus a closed form, or algebraic versus geometric versus counting versus brute-force enumeration.

Then output exactly this line and nothing after it:\\
SCORE: \textlangle integer 0--100\textrangle \quad (100 = certainly the SAME method as A, 50 = unsure, 0 = certainly DIFFERENT methods)

PROBLEM: \{problem\}\\
SOLUTION A: \{modal solution\}\\
SOLUTION B: \{candidate\}
\end{tcolorbox}

\paragraph{Embedding Model Prompt.} We leverage the fact that Qwen3-Embedding-0.6B is instruction-tuned to provide directions aligning with our goal of reasoning diversity.

\begin{tcolorbox}[promptbox, title=Qwen3-Embedding-0.6B Instruction]
Instruct: Represent the core solution strategy and reasoning approach of this text (the sequence of mathematical steps and technique used), not its tone, reasoning narration style, wording, notation, verbosity, or formatting. Two texts should be close only if they use the same strategy.\\
Query: \{solution text\}
\end{tcolorbox}

\paragraph{Method-Diversity Analysis Prompt.} \mbox{}
\begin{tcolorbox}[promptbox, title=Claude Opus 5.5 Diversity Judge Instruction]
You are an expert competition mathematician analysing how solutions to one math problem differ.
You will receive two sets of solutions, Set A (ids A01, A02, ...) and Set B (ids B01, ...). The sets
come from two different sources. They may differ in size. Every
solution reaches the correct final answer. Analyse each set on its own merits. Do not try to make
the sets look similar or different.

METHOD: the core strategy of each solution. Two solutions share a
method if a mathematician would summarise both the same way in one sentence (``complementary
counting'', ``Vieta's formulas'', ``coordinates and direct computation'', ``inscribed-angle chase'').
Different methods rest on a different key insight, reformulation, or theorem.

Within one method, compare solutions pairwise:
- ``variant'': the same key idea, but a mathematically meaningful difference in execution, such as a
  different intermediate quantity or substitution, a different case split, or a reordering of major steps that changes what is computed.
- ``surface'': no mathematical difference, only wording, verbosity, notation, formatting, order of
  presentation, narration, or a final check that recomputes the same thing.

Rules:
- Classify each solution by the approach that actually produced its answer. If it starts one
  approach, abandons it, and finishes another, use the one it finished and set
  abandoned\_attempt = true.
- Use one shared list of methods for both sets: the same method gets the same method\_id wherever it
  appears. A method may appear in only one set.
- Never split methods on variant or surface differences. Do not merge methods merely because they
  share routine algebra. It is common (and valid) for a single method to be used for every completion.
- Assign every solution id to exactly one method.
- Report every pair of solutions that are in the same set AND the same method. Never pair across sets.

Work in this order: summarize each solution in one line, define the methods, assign each solution to
a method, then label the within-set, within-method pairs.
\end{tcolorbox}

\section{Additional Training Dynamics}
\label{app:more_graphs}
\subsection{Additional pass@k}
\begin{figure}[h]
  \centering
  \includegraphics[width=0.9\linewidth]{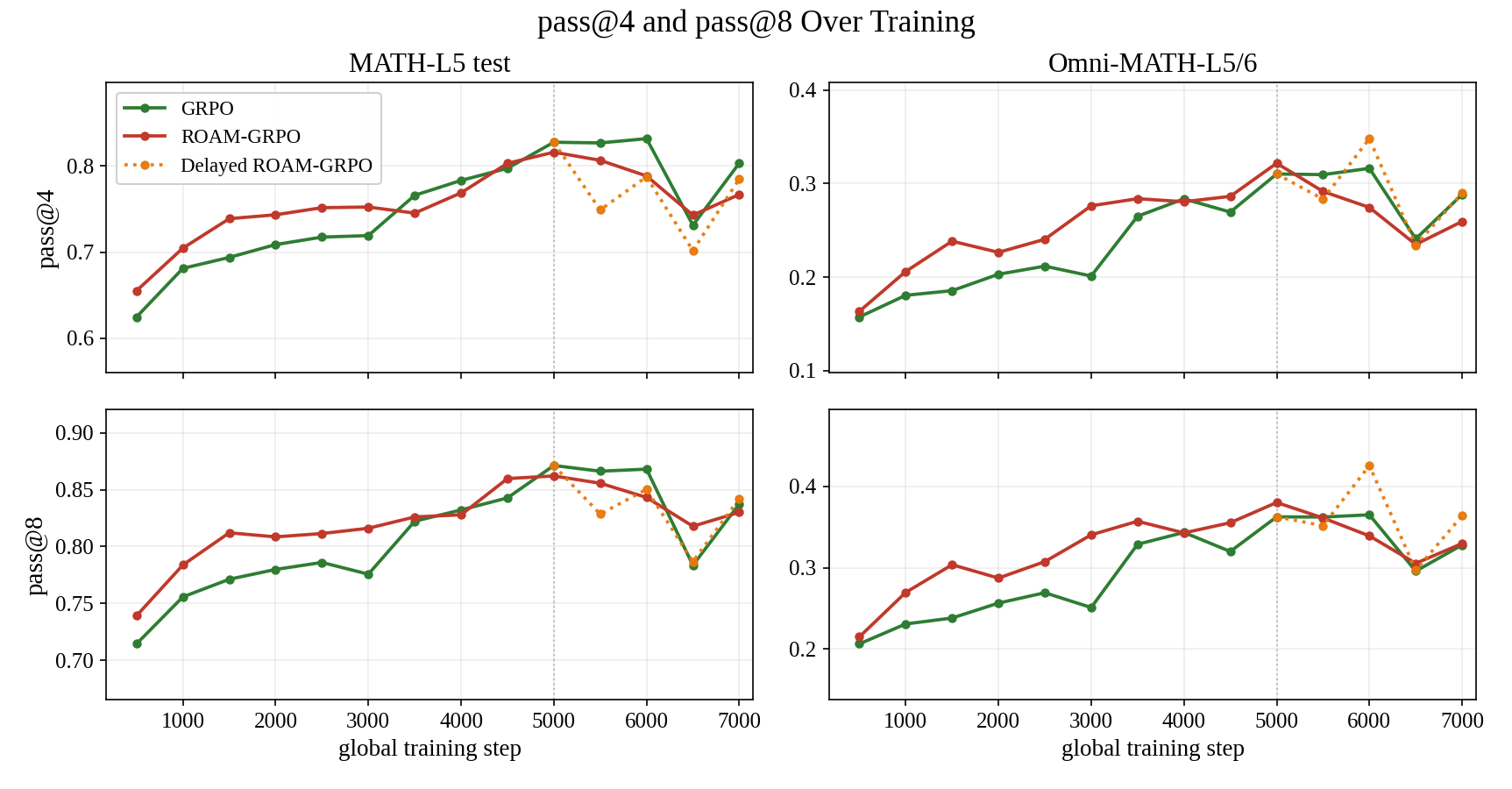}
  \caption{pass@4 and pass@8 over training (MATH-L5, left; Omni-MATH L5/6, right). GRPO, ROAM-GRPO, and delayed ROAM-GRPO (activated at step $5000$, dotted). pass@4 and pass@8 approximately interpolate pass@1 and pass@16: ROAM-GRPO nears or overtakes GRPO in pass@k as k increases. }
  \label{fig:pass_4_8}
\end{figure}

Included here are pass@4 and pass@8 evaluations (Fig. \ref{fig:pass_4_8}) reported along the same training trajectories as in Section \ref{subsec:training}.

Additionally, Table \ref{tab:pass16-sets} provides a finer-grained comparison of pass@16 performance of GRPO and ROAM (taken at their peak checkpoints on each benchmark) by reporting the overlap between the problems correctly solved by each method. On MATH, GRPO and ROAM-GRPO largely overlap in which problems they solve (intersection accuracy is $88.0\%$ compared to ROAM-GRPO's $89.8\%$). On Omni-MATH, the difference is much more substantial: intersection accuracy is $4.8\%$ lower than GRPO, indicating that evaluation is more variable on a held-out difficult test set.

\begin{table}[h]
\caption{Per-problem pass@16 percentage accuracy overlap between GRPO and ROAM at their peak
checkpoints. GRPO $\cap$ ROAM denotes problems solved by both checkpoints, GRPO $\cup$ ROAM denotes problems solved by either. MATH-L5 compares GRPO
(step 5000) vs.\ ROAM-GRPO (step 4500), Omni-MATH compares GRPO (step 5000) vs.\ delayed ROAM-GRPO
(step 6000).}
\centering
\begin{tabular}{lcccc}
\toprule
\textbf{Benchmark} & \textbf{GRPO} & \textbf{ROAM} & \textbf{GRPO $\cap$ ROAM} & \textbf{GRPO $\cup$ ROAM} \\
\midrule
MATH-L5 & 91.0 & 89.8 & 88.0 & 92.8 \\
Omni-MATH-L5/6  & 41.3 & 46.0 & 36.5 & 50.8 \\
\bottomrule
\end{tabular}
\label{tab:pass16-sets}
\end{table}

\subsection{Routing Statistics and Examples}
\begin{figure}[h]
  \centering
  \includegraphics[width=0.75\linewidth]{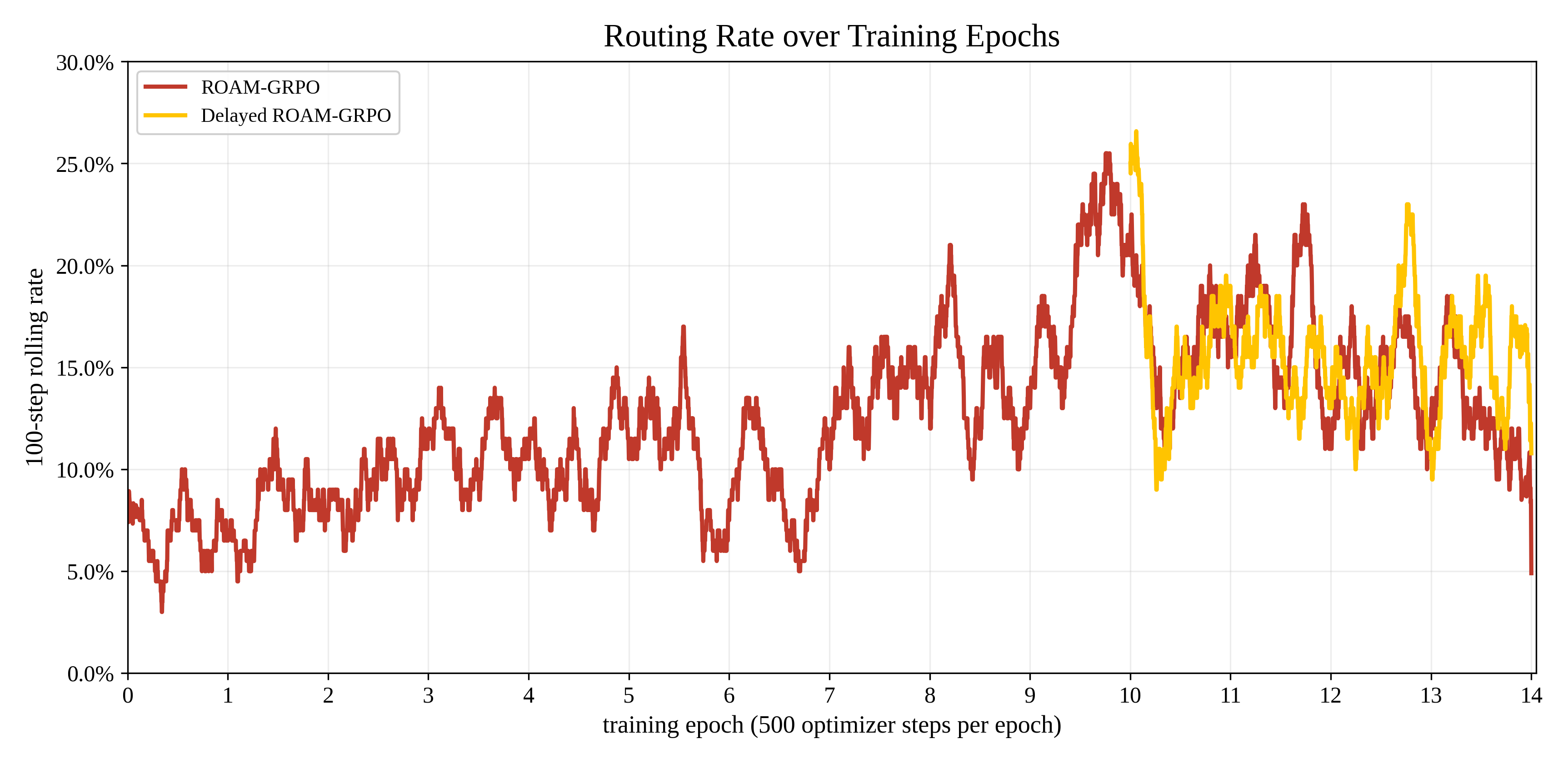}
  \caption{Routing rate of ROAM-GRPO and Delayed ROAM over training, taken over a $100$-step rolling mean. A prompt group is routed when all $G=8$ group completions are correct. Routing slightly increases in frequency across training.}
  \label{fig:routing}
\end{figure}

We provide additional statistics regarding the routing of problems during training (Fig. \ref{fig:routing}), and examples of routed problems that exhibit distinct solutions (abridged to remove unnecessary reasoning chains for readability, Fig. \ref{fig:routing_examples}).

\begin{figure}[h]
\centering
\footnotesize
\begin{tcolorbox}[promptbox, fontupper=\footnotesize, title={Example 1: maximize $fg+gh+hj+fj$ for $\{f,g,h,j\}=\{5,6,7,8\}$ \hfill (answer: 169)}]
\begin{minipage}[t]{0.47\linewidth}
\textbf{Modal solution} (case checking)\\[2pt]
[...] Let's try matching:
Try $f=5, g=8, h=7, j=6$: sum $= 40+56+42+30 = 168$.\\
Try $f=5, g=7, h=8, j=6$: sum $= 35+56+48+30 = 169$.\\
Try $f=5, g=8, h=6, j=7$: sum $= 160$.\\
Try $f=5, g=7, h=6, j=8$: sum $= 155$.\\
So $169$ seems best. \hfill $\boxed{169}$
\end{minipage}\hfill
\begin{minipage}[t]{0.49\linewidth}
\textbf{Accepted off-mode solution} (factorization)\\[2pt]
[\ldots] Let's rearrange:
$fg + gh + hj + fj = g(f+h) + j(f+h) = (g+j)(f+h)$.\\
{[...]}
To maximize product of two sums from a set of 4 numbers, pick largest pairs?\\
1. $\{5,8\},\{6,7\} \to 13 \times 13 = 169$.\\
2. $\{5,7\},\{6,8\} \to 12 \times 14 = 168$.\\
3. $\{5,6\},\{7,8\} \to 11 \times 15 = 165$.\\
The maximum is 169. \hfill $\boxed{169}$
\end{minipage}
\end{tcolorbox}

\begin{tcolorbox}[promptbox, title={Example 2: \# of $a \le 50$ for which $x^2+(2a+1)x+a^2=0$ has two integer roots \hfill (answer: 6)}]
\begin{minipage}[t]{0.47\linewidth}
\textbf{Modal solution} (discriminant)\\[2pt]
[...] From Vieta's formulas: Sum of roots: $(x_1 - x_2)^2 = (x_1+x_2)^2 - 4x_1x_2 = (2a+1)^2 - 4a^2 = 4a+1$.
[...] For integer roots, $4a+1$ must be a perfect square $k^2$.
Since $k^2$ is odd, $k$ odd, let $k = 2m+1$. $4m(m+1) = 4a \Rightarrow a = m(m+1)$.\\
So $m(m+1) \le 50$. So $m \in \{1,\ldots,6\}$.
$a \in \{2, 6, 12, 20, 30, 42\}$. \hfill $\boxed{6}$
\end{minipage}\hfill
\begin{minipage}[t]{0.49\linewidth}
\textbf{Accepted off-mode solution} (consecutive-integer factoring)\\[2pt]
[...] Notice $(x+a)(x+a+1) = x^2 + (2a+1)x + a^2 + a$.\\
So $x^2 + (2a+1)x + a^2 = (x+a)(x+a+1) - a = 0$, i.e.\ $(x+a)(x+a+1) = a$.\\
Since $x$ is an integer, $x+a$ and $x+a+1$ are consecutive integers whose product is $a$.
[...] So $k \in \{1, 2, 3, 4, 5, 6\}$. This gives
$a \in \{2, 6, 12, 20, 30, 42\}$. \hfill $\boxed{6}$
\end{minipage}
\end{tcolorbox}
\caption{Examples of off-mode solutions from ROAM routing during training. Example 1 occurs during ROAM-GRPO training and Example 2 during delayed ROAM-GRPO. Solutions are lightly abridged.}
\label{fig:routing_examples}
\end{figure}

\section{Evaluation Examples}
\label{app:examples}
Here we inspect three additional illustrative comparisons between GRPO and ROAM-GRPO. All recorded completions to these problems are available in the code repository. Abridged completions are provided in the diagrams.

\begin{figure}[h]
  \centering
  \includegraphics[width=0.95\linewidth]{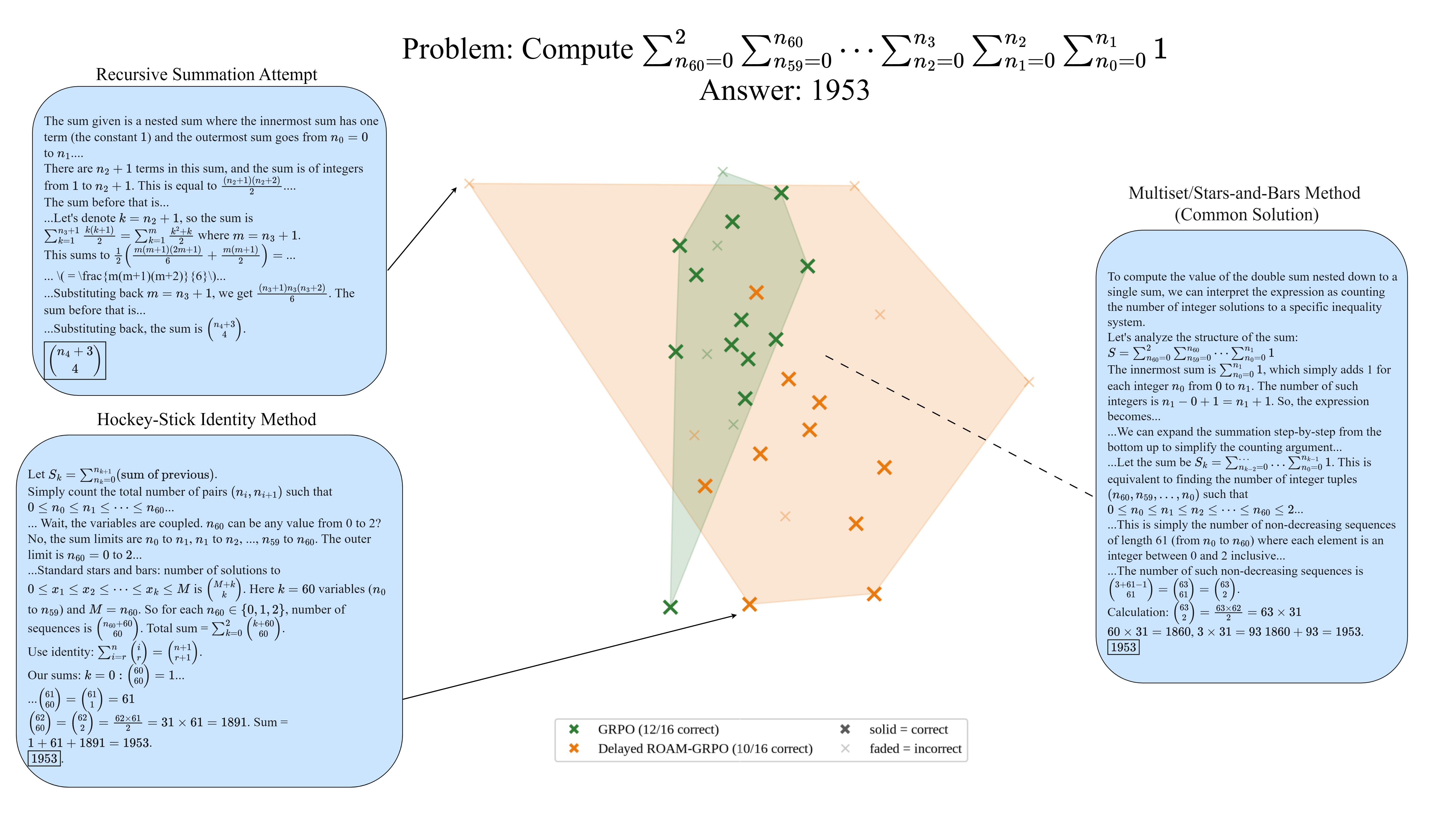}
  \caption{2-D MDS embeddings of attempted solutions on Omni-MATH Eval Problem HMMT\_2/246,
  16 each across GRPO (step 5000, green) and delayed ROAM-GRPO (step 6000, orange), taken at their highest pass@16 checkpoint. 
  All GRPO completions share a common approach (reframing as a combinatorial problem using the stars-and-bars method): 12/16 are correct. Delayed ROAM also mostly employs the common approach, but also adopts an alternative approach (using the hockey-stick identity) twice (succeeds once, fails once). On one occasion it attempts to solve by recursive summation.}
  \label{fig:246}
\end{figure}
\begin{figure}[h]
  \centering
  \includegraphics[width=0.9\linewidth]{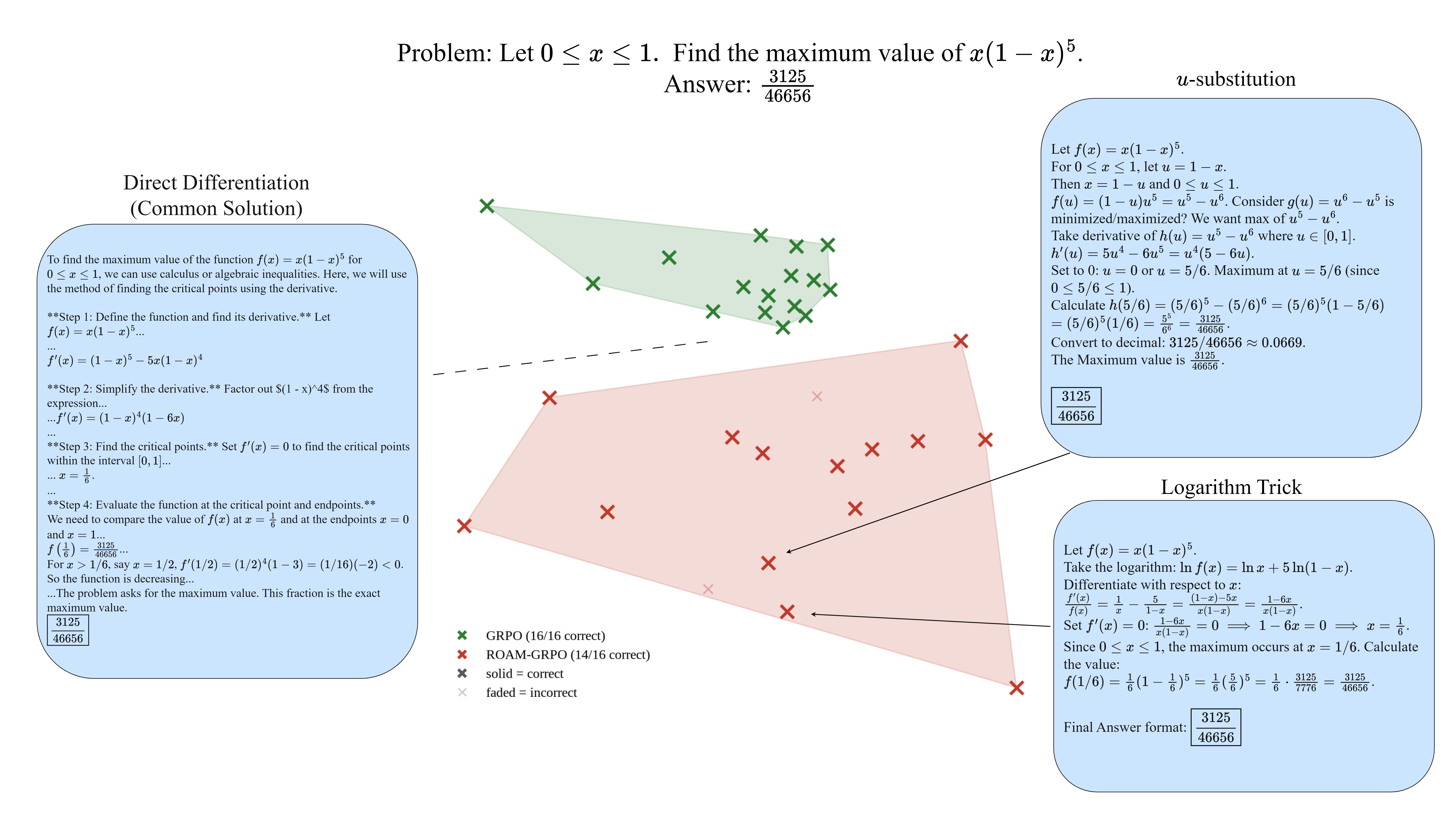}
  \caption{2-D MDS embeddings of attempted solutions on MATH Eval Problem 386,
  16 each across GRPO (step 5000, green) and ROAM-GRPO (step 4500, red), taken at their highest pass@16 checkpoint. 
  All GRPO completions are correct and use the same approach (direct differentiation of the function). 14/16 ROAM-GRPO completions are correct. Most employ the common differentiation approach, but one uses the logarithm trick to simplify differentiation, while another uses $u$-substitution. Both incorrect completions are caused by calculation errors when attempting the common solution. On one occasion, GRPO verifies using the logarithm trick after deriving with the common solution.}
  \label{fig:386}
\end{figure}

\begin{figure}[h]
  \centering
  \includegraphics[width=0.9\linewidth]{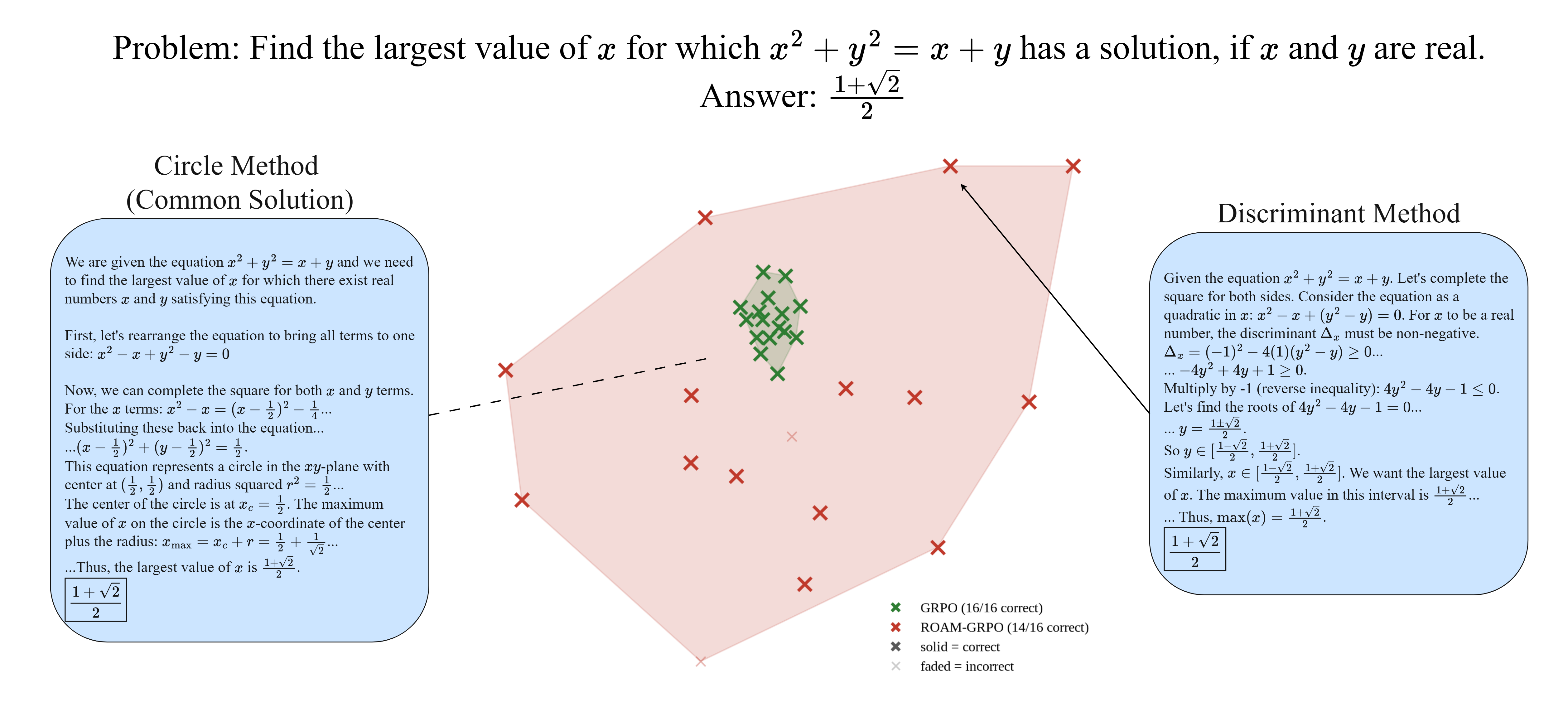}
  \caption{2-D MDS embeddings of attempted solutions on MATH Eval Problem 35, 16 each across GRPO (step 5000, green) and ROAM-GRPO (step 4500, red), taken at their highest pass@16 checkpoint.
  All GRPO completions are correct and use the same approach (viewing geometrically as the function of a circle). 14/16 ROAM-GRPO completions are correct. Most employ the common geometric approach, but one solves the problem purely algebraically through the function discriminant. Both incorrect completions are caused by calculation errors when attempting the common solution.}
  \label{fig:35}
\end{figure}

Figures \ref{fig:246} to \ref{fig:35} all demonstrate GRPO employing a dominant approach while the corresponding ROAM-GRPO variant attempts multiple. Notably, ROAM-GRPO's most common approach often coincides with GRPO's dominant approach, but distinct outliers use computational shortcuts or entirely different methods. These are not always successful, meaning that increased diversity can cost accuracy (shown by 
the recursive summation attempt in Fig. \ref{fig:246}, as well as the relationship between pass@1 and pass@16 in Fig. \ref{fig:training}). 

\clearpage

In Figure \ref{fig:35}, GRPO outputs form a tight cluster enclosed within the ROAM-GRPO hull, despite sharing the same approach as most ROAM-GRPO outputs. This is caused by GRPO yielding highly similar phrasing and syntax in addition to underlying reasoning.

\end{document}